\documentclass[10pt,twocolumn,letterpaper]{article}

\usepackage[pagenumbers]{cvpr} 

\newcommand{\model}{AutoSculpt}
\newcommand{\pSpace}{\vspace{1mm}}

\definecolor{cvprblue}{rgb}{0.21,0.49,0.74}
\usepackage[
    pagebackref,
    breaklinks,
    colorlinks,
    linkcolor=red,      
    urlcolor=blue,      
    citecolor=green,    
    linkbordercolor=red 
]{hyperref}

\usepackage{algorithm}
\usepackage{algorithmicx}
\usepackage{algpseudocode}
\usepackage{multirow}

\def\paperID{10681} 
\def\confName{CVPR}
\def\confYear{2025}

\title{AutoSculpt: A Pattern-based Model Auto-pruning Framework Using Reinforcement Learning and Graph Learning}
\author{
Lixian Jing\\
Ocean University of China\\
{\tt\small jlx@stu.ouc.edu.cn}
\and
Jianpeng Qi\\
Ocean University of China\\
{\tt\small qijianpeng@ouc.edu.cn}
\and
Junyu Dong\\
Ocean University of China\\
{\tt\small dongjunyu@ouc.edu.cn}
\and
Yanwei Yu\\
Ocean University of China\\
{\tt\small yuyanwei@ouc.edu.cn}
}

\begin{document}
\maketitle

\begin{abstract}
As deep neural networks (DNNs) are increasingly deployed on edge devices, optimizing models for constrained computational resources is critical. Existing auto-pruning methods face challenges due to the diversity of DNN models, various operators (e.g., filters), and the difficulty in balancing pruning granularity with model accuracy. To address these limitations, we introduce AutoSculpt, a pattern-based automated pruning framework designed to enhance efficiency and accuracy by leveraging graph learning and deep reinforcement learning (DRL). AutoSculpt automatically identifies and prunes regular patterns within DNN architectures that can be recognized by existing inference engines, enabling runtime acceleration. Three key steps in AutoSculpt include: (1) Constructing DNNs as graphs to encode their topology and parameter dependencies, (2) embedding computationally efficient pruning patterns, and (3) utilizing DRL to iteratively refine auto-pruning strategies until the optimal balance between compression and accuracy is achieved. Experimental results demonstrate the effectiveness of AutoSculpt across various architectures, including ResNet, MobileNet, VGG, and Vision Transformer, achieving pruning rates of up to 90\% and nearly 18\% improvement in FLOPs reduction, outperforming all baselines. The codes can be available at \url{https://anonymous.4open.science/r/AutoSculpt-DDA0}
\end{abstract}    
\section{Introduction}
\label{sec:intro}
Deploying resource-intensive deep neural networks (DNNs) on edge devices, such as mobile phones, robots, and self-driving cars, presents significant challenges due to limited computing resources. This makes model compression, particularly \textit{model pruning}, an essential strategy \cite{survey-dnn-prun, edge-intelligence}. By removing less critical parameters through pruning, DNNs can effectively reduce their computational requirements, enhancing their suitability for resource-constrained scenarios \cite{gnn-rl_yu2022topology, DistrEdge}.

\begin{figure}
    \centering
    \includegraphics[width=1.0\linewidth]{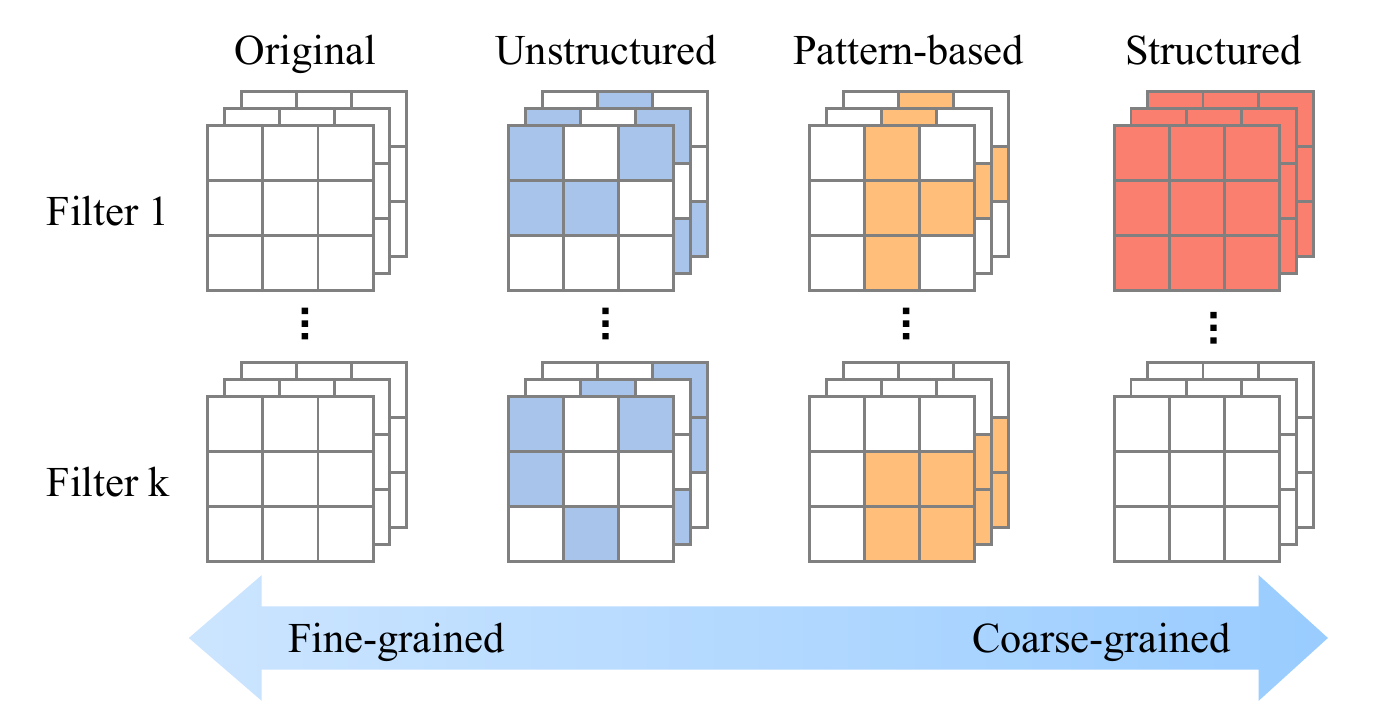}
    \caption{Examples of three different pruning granularity methods on convolutional layers consisting of $k$ filters. Colored blocks represent pruned weights.}
    \label{fig:pruning_granularity}
\end{figure}

Model pruning can be categorized based on the granularity of the process into three types (see \figurename~\ref{fig:pruning_granularity}): \textit{Unstructured pruning} (or fine-grained pruning) \cite{han2015deep, Park*2020Lookahead:, sanh2020movement, louizos2018learning}, \textit{pattern-based pruning} (or semi-structured pruning) \cite{li2021npas, lixiang2024dtmm}, and \textit{structured pruning} (or coarse-grained pruning) \cite{zheng2022model, zhai2023lapp, depgraph_fang2023, li2022revisiting, liu2021group}. For instance, unstructured pruning can remove weights from arbitrary locations in a $3 \times 3$ weight tensor, leading to an irregular distribution of non-zero weights. This irregularity requires specialized software and hardware support for effective acceleration \cite{ma2021non, zhou2021learning}. In contrast, structured pruning removes entire channels, filters (e.g., filter 1 in \figurename~\ref{fig:pruning_granularity}), or layers, preserving a regular network structure that is easier to execute on standard inference engines. However, due to its coarser granularity, structured pruning often struggles to achieve an optimal balance between compression rate and accuracy, potentially due to the loss of critical topology information \cite{depgraph_fang2023, gnn-rl_yu2022topology}.

Pattern-based pruning represents an intermediate approach, focusing on removing specific ``regular structures'' (highlighted in yellow in \figurename~\ref{fig:pruning_granularity}) within the weight tensor that are conducive to hardware efficiency \cite{li2021npas, lixiang2024dtmm}. The development of specialized model compilers, such as TVM \cite{chen2018tvm} and TensorRT \cite{tensorRT21}, has made pattern-based pruning more practical. These compilers can automatically adapt to pruned structures without manual intervention, significantly enhancing runtime acceleration by 
bypassing zeroed regions \cite{zhou2021learning}. Nonetheless, pattern-based pruning techniques are still underexplored, indicating a pressing need for further research.

To search for optimal pruning policies, automated approaches such as AutoML (Automated Machine Learning) \cite{survey-dnn-prun, amc_he2018} have recently gained attention. These methods use graph learning techniques, specifically graph neural networks (GNNs), to represent or embed DNNs. The embeddings are then fed into deep reinforcement learning (DRL) algorithms to find the optimal pruning policy, achieving greater compression efficiency \cite{amc_he2018, ActivationSparsity2020, gnn-rl_yu2022topology}. However, while existing automatic pruning methods have made significant progress, they focus primarily on structured and unstructured pruning, and pattern-based pruning approaches \cite{li2021npas, lixiang2024dtmm, niu2020patdnn, ma2020pconv} are limited and face \textbf{\textit{three main challenges}}:
\textit{(1) They mainly handle convolutional neural networks (CNNs) and lack generalizability to various DNN architectures, especially in representing them, ignoring parameters in architectures like Transformers \cite{vaswani2017attention};
(2) they are predominantly designed for specific operators\footnote{We unify the concepts of convolution (conv), filters, and encoder/decoder blocks under the term ``operator''.} with fixed size (e.g., $3 \times 3$ conv.) in CNNs, which might limit their effectiveness for other shapes;
(3) there is potential to more fully leverage pattern information to improve the effectiveness of the pruning strategies.}

To address these challenges, we propose a universal pattern-based model auto-pruning framework, namely \textit{\model}. We first construct various pretrained DNNs as graphs, and integrate regular patterns that are adaptable to diverse operators into the graph. Then, we utilize a GNN encoder to learn the graph embeddings, and feed these embedings into DRL to automate the search for optimal pruning patterns.
To validate the feasibility and effectiveness of \model, we conduct experimental evaluations across several datasets and serveral DNNs with different architecture, including ResNet, MobileNet, VGG, and Vision Transformer. Our results are compared against state-of-the-art (SOTA) methods, demonstrating that our framework achieves competitive performance. 
Specifically, we achieve a pruning ratio that exceeds all baseline methods, reaching up to 90\%, while reducing FLOPs by nearly 18\% compared to the latest auto-pruning method, and achieving SOTA performance levels.

In summary, our contributions are as follows:
\begin{itemize}[leftmargin=2em]
    \item We propose a universal pattern-based auto-pruning framework, \model, that supports a variety of DNN models, integrating GNN and DRL to effectively leverage DNN topology, weight parameter information, and regular zero-region patterns in pruning decisions.
    \item We combine pattern representation with DRL to achieve automated pruning and designed a reward function suitable for pattern pruning, effectively training agents to complete the pattern policy search.
    \item We validate \model\ through experiments on various popular DNNs, achieving SOTA results.
\end{itemize}

\section{Related Work}
\label{sec:relatedwork}
{\bf Pruning Granularities.} Model pruning has emerged as a popular research direction to enhance inference performance on resource-constrained devices. 
Generally, it can be categorized into unstructured, structured, and semi-structured pruning \cite{survey-dnn-prun}. 
Unstructured pruning \cite{han2015deep, sanh2020movement, louizos2018learning, Park*2020Lookahead:} removes weights arbitrarily across the network, theoretically achieving the highest pruning rates without altering the network’s overall structure. However, the resulting irregularity and sparsity of the compressed weight tensors make hardware acceleration challenging. Consequently, most researchers have shifted their focus to structured pruning methods \cite{he2023structured}. Structured pruning \cite{amc_he2018, gnn-rl_yu2022topology, depgraph_fang2023, greg_wang2021neural, abp_tian2021adding, dlrfc_he2022filter, hrank_lin2020, chip_sui2021, mfp_he2022filter, soks_liu2022, dncp_zheng2022model, nm_kim2020neuron} eliminates entire filters, channels, or layers, leading to more regular and hardware-friendly network architectures. Mainstream approaches in structured pruning include applying sparse regularization techniques to model parameters during training, such as LASSO \cite{wen2016learning} and ADMM \cite{li2019compressing}; dynamically adding masks to weights during training and inference for pruning (also known as soft pruning) \cite{he2018soft, ddg_li2021dynamic, smcp_humble2022soft}; and utilizing mathematical techniques like second-order Taylor approximation \cite{wang2019eigendamage} and Variational Bayesian methods \cite{zhou2019accelerate} for pruning solutions. However, the granularity of structured pruning can be too coarse, potentially resulting in the removal of important parameters. Semi-structured (or pattern-based) pruning \cite{li2021npas, ma2020pconv, niu2020patdnn} seeks suitable ``regular patterns'' to be removed on weight matrices. Models pruned using pattern-based methods exhibit more regular zero-shaped regions, thereby achieving inference acceleration while maintaining accuracy.

\begin{figure*}[htbp]
  \centering
  \includegraphics[width=1.0\linewidth]{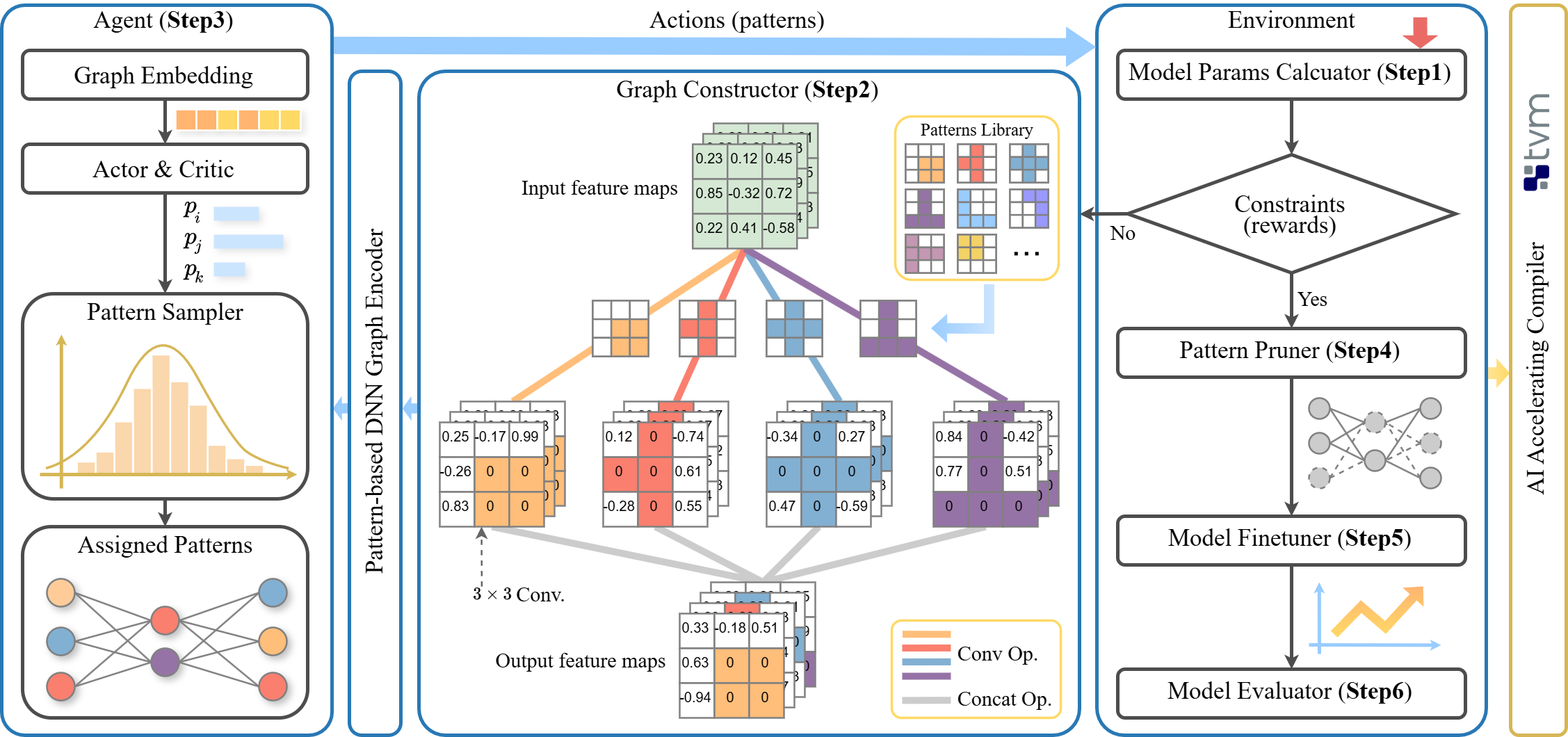}
  \caption{An overview of our \model\ framework.}
  \label{fig:modelfig}
  \vspace{-5mm}
\end{figure*}

\pSpace
\noindent{\bf Automatic Pruning Methods.} The architectures of modern DNNs are becoming increasingly complex and contain rich topology information, making it difficult to achieve satisfactory model compression results through manual heuristic methods \cite{ma2020pconv, niu2020patdnn, gnn-rl_yu2022topology, lixiang2024dtmm}. Consequently, AutoML united GNN and DRL becomes mainstream.
MetaPruning \cite{liu2019metapruning} pre-trains a meta-network to predict the weights of the target DNN and then uses an evolutionary process to search for the best-performing pruned network. 
AMC \cite{amc_he2018} is the first to introduce DRL into structured pruning for CNN, performing model compression by searching redundant channels. 
AGMC \cite{agmc_yu2021auto} and GNN-RL \cite{gnn-rl_yu2022topology} model CNN as graph, obtaining graph representations (or embeddings) through GNN and using DRL to pruning entire convolution or pooling layers. 
DTMM \cite{lixiang2024dtmm} proposed a pattern called ``filterlet'' as the pruning unit and designed a convolution operator to optimize the pruning strategy with the help of the pruning strategy scheduler. 
NPAS \cite{li2021npas} proposes an automated pattern pruning and architecture search framework with compiler awareness.

However, existing work mainly focuses on coarse blocks (or layers) or less diverse types of DNNs. The pruning granularity in these approaches may limit their applicability to varous DNNs such as Transformer. There is still insufficient research on how to embed regular patterns into various DNN architectures, utilize DRL to identify optimal pruning strategies, and ultimately boost inference performance through automated techniques and regular patterns.
\section{Method}
\label{sec:method}

We first introduce the \model\ framework in Section \ref{sec:overview-framework}, and then elaborate on its two key components: The pattern-based graph constructor and encoder in Section \ref{sect:graph-constructor} and the DRL-based pruning strategy search in Section \ref{sect:drl-pruning}.


\subsection{Overview of \model}\label{sec:overview-framework}

AutoSculpt is a multi-step framework for pruning DNN models, as shown in \figurename~\ref{fig:modelfig}. First, we calculate the compression parameters (e.g. inference accuracy, FLOPs and pruning ratio) of the DNN models to determine if they meet the constraints $\mathcal{C}$ (Step 1). If the constraints are unmet, the \textit{Graph Constructor} extracts topological features from the model and assigns pruning patterns to build the graph (Step 2). Next, we use a GNN, specifically the Graph Attention Network (GAT), to encode the graph and obtain the embeddings. Then, the \textit{Agent}, calculates the probability distribution for each pattern based on graph embeddings and selects suitable pruning patterns to optimize the model using \textit{Pattern Sampler} while interacting iteratively with the environment (Step 3). Once constraints are satisfied, we can obtain the pruned DNN through \textit{Pattern Pruner} (Step 4), followed by fine-tuning to restore accuracy (Step 5). Finally, the pruned model can be deployed using an AI compilation framework (Step 6). 

\subsection{DNN-Graph Constructor and Encoder}\label{sect:graph-constructor}
    
    


A general graph $\mathcal{G}$ constructed from a DNN $\mathcal{W}$ can be represented as $\mathcal{G} = \{ \mathcal{V}, \mathcal{E} \}$, where $\mathcal{V} = \{ \mathcal{N}_{i}, \mathcal{N}_{k}, \mathcal{N}_{o} \}$ is the set of nodes and $\mathcal{E} = \{ \mathcal{E}_{i}, \mathcal{E}_{o} \}$ is the set of edges. The subsets $\mathcal{N}_{k}$ represent nodes associated with weight tensors, $\mathcal{N}_{i}$ is the input tensor node set, and $\mathcal{N}_{o}$ is the output node set. The subsets $\mathcal{E}_{i}$ and $\mathcal{E}_{o}$ correspond to the edges linked to $\mathcal{V}$. The nodes in different layers of the DNN can be further divided into finer-grained sets. For example, the nodes in the $l$-th layer are contained in $\mathcal{N}_{k}^{l} \subset \mathcal{N}_{k}$. Next, we describe the graph constructors of two typical DNN architectures\footnote{In this paper, we focus on CNN and Transformer models, but other types of DNN can also be applied.} : CNN and Transformer.

\pSpace
\noindent{\bf CNN-Graph Constructor.}
Given a general CNN $\mathcal{W}$, we use a set of weights $\{ \mathbf{W}_i ^{l}\in \mathbb{R}^{c \times k \times k}, 0 \le l \le m, 0 \le i \le n_l \}$ to parameterize the architechture of CNN, where $\mathbf{W}_i ^{l}$ represents the weight tensor with $c$ channels of $i$-th $k \times k$ kernel of $l$-th layer, and $m$ is the number of convolutional layers, $n_l$ is the number of convolutional kernels in $l$-th layer. And the related input feature maps and output feature maps (with $w \times h$ size) of $l$-th layer are $\mathbf{X}^{l}_{in} \in \mathbb{R}^{c \times w \times h}$ and $\mathbf{X}^{l}_{out} \in \mathbb{R}^{m \times w \times h}$, respectively.

\begin{figure}
    \centering
        \includegraphics[width=1.0\linewidth]{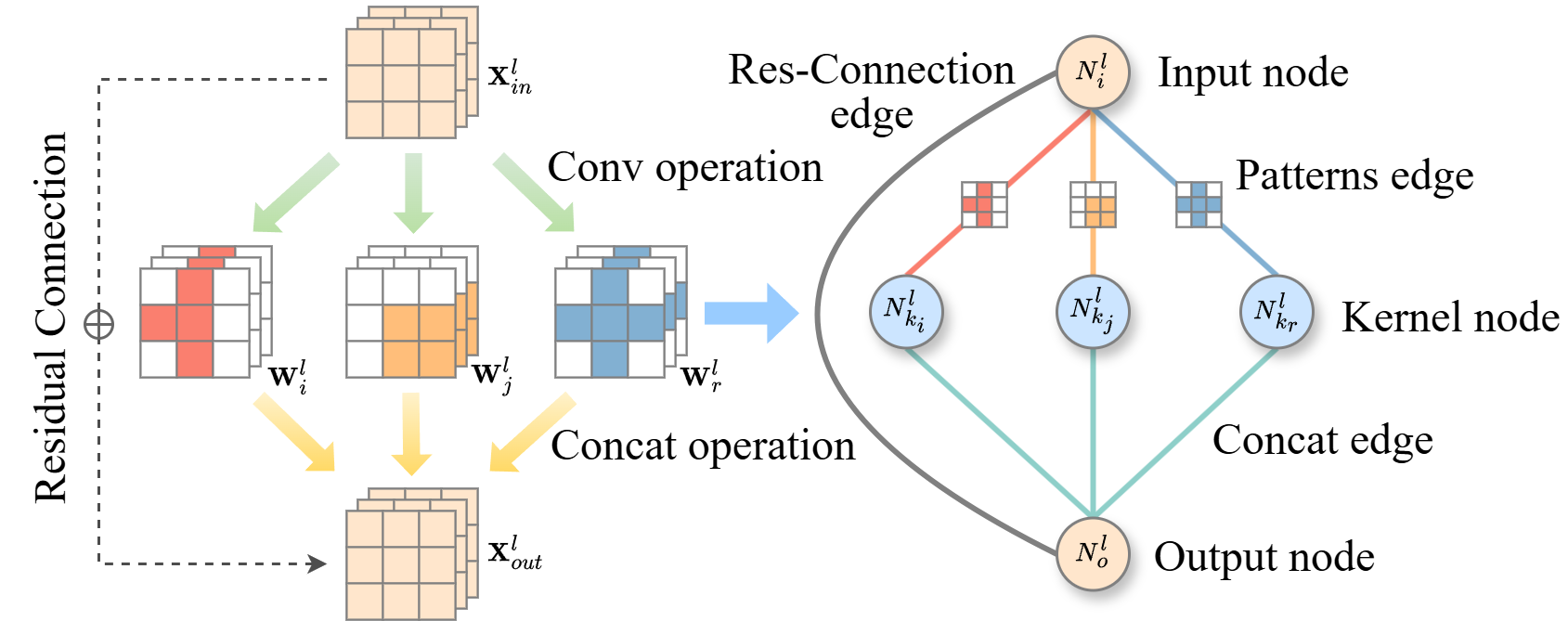}
        \caption{An Example of Graph Construction for CNN.}
        \label{fig:cnn_construct}
\vspace{-4mm}
\end{figure}
Then, the specific process of graph construction is as follows: First, map the weight set of the $l$-th layer $\{ \mathbf{W}_i ^{l}\in \mathbb{R}^{c \times k \times k}, 0 \le i \le n^l \}$ to a set of nodes $\mathcal{N}_{k}^{l} \subset \mathcal{V}$, where a node $N_{k_i}^{l} \in \mathcal{N}_{k}^{l}$ represents the weight of the kernel $\mathbf{W}_i ^{l}$. At the same time, map the input and output feature maps of the $l$-th layer to the nodes $N_{i}^{l} \in \mathcal{N}_{i}^{l}$ and $N_{o}^{l} \in \mathcal{N}_{o}^{l}$, respectively. The dependencies of the convolution operation are then mapped to the connection relationships between the node set $\mathcal{N}_{k}^{l}$ and the node sets $\mathcal{N}_{i}^{l}$ and $\mathcal{N}_{o}^{l}$, forming the sets of edges $\mathcal{E}_{i}^{l}, \mathcal{E}_{o}^{l} \subset \mathcal{E}^{l}$ in the graph $\mathcal{G}$. After completing the above steps, we obtain the graph $\mathcal{G}$, as shown in \figurename~\ref{fig:cnn_construct}. 
Since the parameters of the CNN model are primarily located in the convolutional layers, and the final fully connected layer connects to the output layer, our focus during graph construction is on the convolutional layers.

After the basic structure of the graph $\mathcal{G}$ is constructed, we integrate the pruning pattern $\mathcal{P}$ (shown as \textit{Patterns Library} derived from Agent in \figurename~\ref{fig:modelfig}) into $\mathcal{G}$. Considering that integrating key information into edge feature embeddings yields better results than integrating it into node feature embeddings ~\cite{qin2023graph}, we adapt the following integration scheme: The pattern $\mathcal{P}$ is fused into the edge feature embeddings of the edge set $\mathcal{E}_{i}$, while the weights of the CNN corresponding to the node set $\mathcal{N}_{k}$ are integrated into their node feature embeddings. In addition, the node feature embeddings in the node sets $\mathcal{N}_{i}$ and $\mathcal{N}_{o}$, as well as the edge feature embeddings in the edge set $\mathcal{E}_{o}$, are assigned using random initialization. In short, we denote this process as \cref{eq:graphconstructor}:
\vspace{-1.5mm}
\begin{equation}
    \mathcal{G} = GraphConstructor(\mathcal{W}, \mathcal{P}).
    \label{eq:graphconstructor}
\vspace{-1.5mm}
\end{equation}

Additionally, in CNN architectures like ResNet, there are residual connections.
These residual connections establish additional pathways between layers, allowing the flow of information across layers without being disrupted by pruning. As a result, when building the graph, we represent the residual connections as additional edges that link nodes corresponding to the layers involved in the residual paths. This ensures that the graph $\mathcal{G}$ accurately reflects the full structure of the CNN, including these crucial connections, which are essential to maintain the performance of the CNN model during pruning.

\pSpace
\noindent{\bf Transformer-Graph Constructor.} 
To model the Transformer encoder $\mathcal{W}^{'} \subset \mathcal{W}$, we assume the input embedding of $l$-th encoder is $\mathbf{X}_{in}^l \in \mathbb{R}^{T \times d}$, where $T$ means the input sequence consists of \(T\) tokens and \(d\) is the embedding dimension. Then, we have \(\mathbf{Q}^l = \mathbf{X}_{in}^l \mathbf{W}_Q^l\), \(\mathbf{K}^l = \mathbf{X}_{in}^l \mathbf{W}_K^l\), and \(\mathbf{V}^l = \mathbf{X}_{in}^l \mathbf{W}_V^l\), where \(\mathbf{W}_Q^l, \mathbf{W}_K^l, \mathbf{W}_V^l \in \mathbb{R}^{d \times d_k}\) are learnable weight matrices for Querys, Keys, and Values, and \(d_k\) is the dimension of each. The attention score is shown as $\mathbf{X}_{\alpha}$ and the output is $\mathbf{X}_{out}^l \in \mathbb{R}^{T \times d}$.

\begin{figure}
    \centering
        \includegraphics[width=1.0\linewidth]{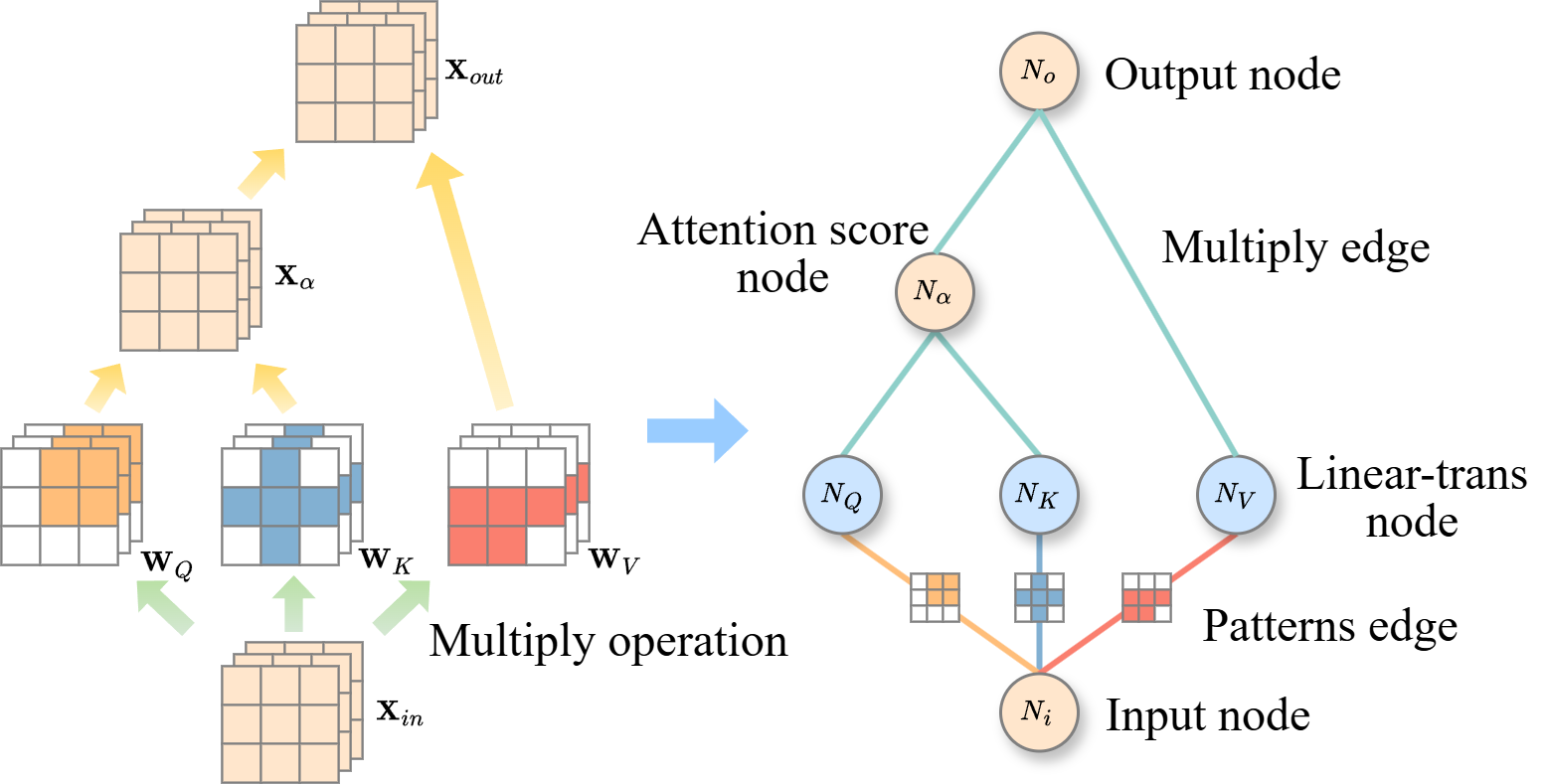}
        \caption{An Example of Graph Construction for Transformer.}
        \label{fig:transformer_construct}
\vspace{-4mm}
\end{figure}

Although the Transformer model structure differs significantly from CNNs, a graph can also be constructed for it to enable pattern pruning. The method of constructing the graph $\mathcal{G}$ for the attention module in a Transformer model is illustrated in \figurename~\ref{fig:transformer_construct}.
For the $l$-th encoder, we map $\mathbf{W}_Q^l, \mathbf{W}_K^l, \mathbf{W}_V^l$ to the nodes $N_Q^l, N_K^l, N_V^l \in \mathcal{N}_{k}^{l}$, respectively. Similarly, the inputs and outputs of the $l$-th encoder are maped to $N_{i}^{l} \in \mathcal{N}_{i}^{l} \subset \mathcal{V}$ and $N_{o}^{l} \in \mathcal{N}_{o}^{l} \subset \mathcal{V}$, respectively. Then, the nodes set $\mathcal{V}$ together with the related edges set $\mathcal{E}$ constitutes the target graph $\mathcal{G}$. 
Additionally, since the Transformer model contains multiple encoder and decoder blocks, each of which includes a feedforward network (MLP) that holds a substantial portion of the model's parameters, this also needs to be considered when constructing the graph. In this case, nodes can represent both the attention mechanism components (such as the Query, Key, and Value matrices) and the MLP layers, while the edges capture dependencies between these components within each encoder and decoder block. 

Note that in this version of \model, we manually define up to 10 pruning patterns in the \textit{Patterns Library} for the Agent, as illustrated in \figurename~\ref{sec:overview-framework}, ensuring that the pattern shapes are consistent with the DNN’s weight tensor shape. Expanding the range of patterns is planned as future work and is beyond the scope of this paper.

\pSpace
\noindent{\bf Pattern-based DNN-Graph Encoder.} After the graph construction is completed, we need to encode the graph $\mathcal{G}$ and extract its representative embeddings $\mathbf{g}$, denoted as \cref{eq:graphencoder}. As an effective graph encoder, GAT introduces an attention mechanism between nodes, allowing us to capture the inner dependencies that exist in the constructed DNN graph. 
In this paper, we use an improved version of GAT, GATv2 \cite{brody2022how}, as the graph encoder. 

\vspace{-2mm}
\begin{equation}
    \mathbf{g} = GraphEncoder(\mathcal{G})
    \label{eq:graphencoder}
\vspace{-2mm}
\end{equation}

In detail, GATv2 introduces a dynamic attention mechanism that can better aggregate information between nodes. For example, to obtain the feature embedding $\mathbf{h}_k^{\prime}$ of a node $N_{k}$ after aggregating information, it can be calculated using \cref{eq:aggregate}:

\begin{equation}
    \mathbf{h}_k^{\prime} = \sigma \left( \sum_{j \in \mathcal{N}_i, \mathcal{N}_o} \alpha_{k,j} \cdot \mathbf{W}_j \mathbf{h}_j \right),
    \label{eq:aggregate}
\end{equation}
\noindent where $\mathbf{h}_k^{\prime}$ is the feature embedding of node $N_{k}$ after information aggregation, the attention coefficient $\alpha_{k,j}$ of $N_{k}$ and $N_{j}$ can be calculated using \cref{eq:alpha_kj}, and $\mathbf{W}_j$ is the weight matrix. $\mathbf{h}_j$ is the feature embedding of node $N_{k}$'s neighbor $N_{j}$.

\begin{equation}
    \alpha_{k,j} =
    \frac{
        \exp \left( \mathbf{a}^\top g(\mathbf{h}_k, \mathbf{h}_j) \right)
    }{
        \sum_{l \in \mathcal{N}_i, \mathcal{N}_o} \exp \left( \mathbf{a}^\top g(\mathbf{h}_k, \mathbf{h}_l) \right)
    }
    \label{eq:alpha_kj}
\end{equation}

\begin{equation}
    g(\mathbf{h}_i, \mathbf{h}_j) =
        \mathrm{LeakyReLU} \left( \mathbf{W}_s \mathbf{h}_i + \mathbf{W}_t \mathbf{h}_j + \mathbf{W}_e \mathbf{e}_{i, j} \right)
    \label{eq:gh_ij}
\end{equation}

In \cref{eq:alpha_kj,eq:gh_ij}, the scoring function $g(\mathbf{h}_i, \mathbf{h}_j)$ is used to calculate the score between node $N_{i}$ and node $N_{j}$, representing the importance of neighbor $N_{j}$ to $N_{i}$. $\mathbf{a}$, $\mathbf{W}_s$, $\mathbf{W}_t$, and $\mathbf{W}_e$ are learnable parameters. $\mathbf{e}_{i, j}$ represents the edge feature embedding between node $N_{i}$ and node $N_{j}$ containing pattern information. 

After the node feature embedings are aggregated, they already contain rich DNN topology information and pattern distribution information. 
Next, global pooling is applied to extract the graph embedding $\mathbf{g}$ from the node feature embeddings. Thus, \cref{eq:graphencoder} can be further elaborated as \cref{eq:graph_embedding}:

\begin{equation}
    \mathbf{g} = \frac{1}{||\mathcal{V}||} \sum_{i=1}^{||\mathcal{V}||} \mathbf{h}_{i}.
    \label{eq:graph_embedding}
\end{equation}

\subsection{DRL-based Pruning Strategy}\label{sect:drl-pruning}
We leverage DRL to find the optimal pruning ratios efficiently. In the following, we describe the details of DRL-based pruning strategy.


\pSpace
\noindent{\bf Environment States $\mathcal{S}$.} Our training objective is to combine the model parameters of the DNN with topological structure information to determine the pruning patterns for its operators. Therefore, we pass the graph embedding $\mathbf{g}$ obtained from the graph encoder as the environment states (\cref{eq:graph_embedding}). Since the pruning pattern applied to the DNN changes after each iteration, the graph $\mathcal{G}$ needs to be reconstructed to update the DNN representation.

\pSpace
\noindent{\bf Action Space $\mathcal{F}$.} The direct output obtained by Agent is the probability distribution of all predefined patterns, making its action space $\mathcal{F}$ continuous: $\mathcal{F} = [a_1, a_2, ..., a_n]$, where $n$ is the number of predefined patterns. The actions can be calculated using \cref{eq:action}, and then sample target patterns $\mathcal{P}^{'}$ for the DNN we prune (represented by \cref{eq:patternsampler}).

\begin{equation}
    \mathcal{F} = Tanh(MLP(\mathbf{g}))
    \label{eq:action}
\end{equation}

\vspace{-2mm}
\begin{equation}
    \mathcal{P}^{'} = PatternSampler(\mathcal{F})
    \label{eq:patternsampler}
\end{equation}
\vspace{-2mm}

\noindent{\bf Reward $R$.} The design of the reward function is crucial for training in DRL. Generally, a higher pruning rate results in a more severe decline in model inference accuracy \cite{survey-dnn-prun}. As training progresses, the DRL Agent tends to favor lower pruning rates to achieve better inference accuracy. Therefore, we design the reward function by combining model compression metrics (using FLOPs as an example, though other metrics like Multiply–Accumulate operations (MACs) can be similarly applied) and model inference accuracy as follows:

\begin{equation}
    R = \alpha FLOPs + (1 - \alpha) Acc,
    \label{eq:reward}
\end{equation}

\noindent where $\alpha$ is a learnable parameter. When $\alpha$ is larger, Agent is encouraged to adopt a more aggressive compression strategy, prioritizing model compression over accuracy. Conversely, when $\alpha$ is smaller, Agent is incentivized to take a more conservative approach. 

After that, to train the Agent, we adapt the PPO-Clip algorithm \cite{schulman2017proximal}.



\pSpace
\noindent{\bf Pattern Pruner.} When meeting the constraints $\mathcal{C}$, we set the corresponding position weights of DNN to zero according to the assigned patterns, which can be represented by \cref{eq:patterpruner}:
\vspace{-2mm}
\begin{equation}
    \mathcal{W}_p = PatternPruner(\mathcal{W}, \mathcal{P}).
    \label{eq:patterpruner}
\end{equation}

\begin{algorithm}
\small
\caption{Pattern Pruning Process}
\label{pruning_flow}
{\bf Input:} DNN $\mathcal{W}$, Constraint set $\mathcal{C}$ \\
{\bf Output:} Pruned DNN $\mathcal{W}_p$
\begin{algorithmic}[1]
\State Initialize DRL environment $\mathcal{S} \gets \mathcal{W}$
\State Initialize DRL Agent with patterns $\mathcal{P}$
\State Initialize ReplayBuffer $\mathcal{B}$
\For{$i = \{1, 2, \dots, ||Episode|| \}$}
    \State Assign patterns $\mathcal{P} \to \mathcal{W}$
    \Repeat
        \State Construct Graph $\mathcal{G}$ by \cref{eq:graphconstructor}
        \State Calculate embedding $\mathbf{g}$ by \cref{eq:graphencoder}
        \State Calculate pattern probs $\mathcal{F}$ by \cref{eq:action}
        \State Regenerated Patterns $\mathcal{P}^{'}$ by \cref{eq:patternsampler}
        \State Assign patterns $\mathcal{P}^{'} \to \mathcal{W}$
        \State Calculate reward $R$ by \cref{eq:reward} 
        \State $\mathcal{B} \gets \mathcal{G}, \mathcal{F}, R$;
        $\mathcal{P} \gets \mathcal{P}^{'}$
    \Until{$FLOPs, Acc$ satisfied $\mathcal{C}$}
    \State Prune $\mathcal{W}$ by \cref{eq:patterpruner}
    \If{$i ==$ UpdateEpisode $t$}
        \State Calculate discounted rewards $Dr$
        \State Update Agent using PPO-Clip
        \State Set $\mathcal{B} \to \varnothing$
    \EndIf
\EndFor
\end{algorithmic}
\end{algorithm}
\noindent{\bf Algorithm.} Algorithm \ref{pruning_flow} summarizes the pattern pruning process. Lines 1-3 initialize tasks, including putting DNN $\mathcal{W}$ into environment $\mathcal{S}$, initializing patterns $\mathcal{P}$ for Agent and initializing replaybuffer $\mathcal{B}$ for storing historical training data. The following lines 4-23 describe the pattern pruning process (as depicted in Section \ref{sec:overview-framework} Steps 1-4).


\section{Experiments}

\subsection{Settings}

\noindent{\bf Datasets and Platform Settings.} We conduct experimental validation on models in the image classification domain. The datasets include CIFAR-10/100 \cite{krizhevsky2009learning} and ImageNet-1K (ILSVRC-2012) \cite{deng2009imagenet}. During the pruning and fine-tuning processes, we utilize all training data, randomly sampling 50\% of the data from the test set for validation. The experiments are carried out on a hardware platform equipped with an AMD EPYC 7402 processor and four RTX 4090 GPUs. Due to the large size of the ImageNet dataset, we implement multi-GPU parallel training.

\pSpace
\noindent{\bf Baselines.} 
To ensure fairness and authority, we directly cite validated results from the original work, and the baselines compared are:
\begin{itemize}[leftmargin=2em]
    \item Regularization-based methods: ABP~\cite{abp_tian2021adding}, SOKS~\cite{soks_liu2022}, SCP~\cite{scp_kang2020operation}, GREG~\cite{greg_wang2021neural}.
    \item Dynamic pruning methods: DDG~\cite{ddg_li2021dynamic}, SMCP~\cite{smcp_humble2022soft}, DRLP~\cite{drlp_liu2021joint}, CP-ViT~\cite{cp_vit_song2022}.
    \item Reinforcement learning-based methods: AGMC~\cite{agmc_yu2021auto}, GNN-RL~\cite{gnn-rl_yu2022topology}, AMC~\cite{amc_he2018}.
    \item Second-order approximation methods: SOSP~\cite{sosp_nonnenmacher2022}, GFP~\cite{gfp_liu2021group}.
    \item Activation-based methods: DLRFC~\cite{dlrfc_he2022filter}, Hrank~\cite{hrank_lin2020}, CHIP~\cite{chip_sui2021}.
    \item Gradient-based methods: MFP~\cite{mfp_he2022filter}, DNCP~\cite{dncp_zheng2022model}.
    \item Other pruning methods: DepGraph~\cite{depgraph_fang2023}, ProsPr~\cite{prospr_alizadeh2022prospect}, 
    RollBack~\cite{rollback_fan2022bayesian}, NM~\cite{nm_kim2020neuron}, CC~\cite{cc_li2021towards}, NPPM~\cite{nppm_gao2021network}, MDP~\cite{mdp_guo2020multi}.
\end{itemize}

\pSpace
\noindent{\bf DNN Settings.} We conducte experiments on various DNNs with different architectures. For ResNet-32/56/110 trained on CIFAR-10 and VGG-19 trained on CIFAR-100, we use self-pretrained parameters for pruning. For MobileNet-v1/v2, ResNet-50, VGG-16 and ViT-B/16 trained on ImageNet, we utilize the pretrained parameters built into PyTorch for pruning. Considering the complexity of optimizing and deploying pruned models, we share the pruning patterns for the residual connection layers when pruning the ResNet models with residual connections. Additionally, we ignore the bias terms in the model computations for all pruned models.

\pSpace
\noindent{\bf Pruning Process Settings.} During the model pruning process, the initial feature embedding size for the nodes in the constructed graph of the DNN is set to $32$ (with the same size for edge feature embeddings). After information aggregation, the feature embedding size increases to $64$, and the overall graph feature embedding size is $256$. 
When using the PPO algorithm for policy updates of the Agent, we employ the Adam optimizer to optimize both the Actor and the Critic. The learning rate for the Actor is set to $3 \times 10^{-3}$, while the learning rate for the Critic is set to $1 \times 10^{-3}$. The discount factor $\gamma$ for controlling the reward attention is set to $0.9$, and the PPO clipping range $\epsilon$ is set to $0.2$. The replay buffer size is set to $32$, and policy updates occur once the replay buffer is full, with a total of $15$ update iterations.

\pSpace
\noindent{\bf Fine-tuning Settings.} During the fine-tuning of the pruned models, we use the SGD optimizer with the following settings: $momentum = 0.9$, $weight\ decay=4 \times 10^{-5}$, and the initial learning rate set to $3 \times 10^{-2}$. 
In the training process, we update the learning rate using a multi-step decay approach, with the decay multiplicative factor $\delta = 0.1$. For models trained on CIFAR-10/100, the milestones are set to \([30, 50, 70, 80, 90]\), totaling $100$ training epochs. For models trained on ImageNet, the milestones are set to \([25, 35, 45, 50]\), totaling $60$ training epochs. 

\begin{table}
    \centering
    \caption{Pruning results on CIFAR-10/100. P.Acc. shows the inference accuracy of pruned DNN, $\Delta$Acc. represents the accuracy difference between the pruned DNN and the original DNN, and FLOPs $\downarrow$ denotes the FLOPs reduction.}
    \setlength{\tabcolsep}{2.5mm}  
    \vspace{-6pt}
    \footnotesize
    \begin{tabular}{c|c|c|c|c}
    \toprule
    \textbf{DNN} & \textbf{Method} & \textbf{P. Acc.} & \textbf{$\Delta$ Acc.} & \textbf{FLOPs $\downarrow$} \\
    \midrule
    \multirow{7}*{ResNet-32} & DDG & 93.21 & -0.01 & 43.40 \\
        & ABP & 92.55 & -0.08 & 46.30 \\
        & SOKS & 92.44 & -0.38 & 46.85 \\
        & AGMC & 90.96 & -1.67 & 50.00 \\
        & GNN-RL & 92.58 & -0.05 & 51.00 \\
        & SOSP & 95.06 & -0.24 & 67.36 \\ \cline{2-5}
        & \textbf{Ours} & 92.16 & -0.20 & \textbf{70.00} \\
    \midrule
    \multirow{9}*{ResNet-56} & MDP & 94.29 & +0.55 & 45.11 \\
        & ABP & 92.55 & -0.08 & 46.30 \\
        & AGMC & 92.76 & -0.63 & 50.00 \\
        & NPPM & 93.40 & +0.36 & 50.00 \\
        & DLRFC & 93.57 & -0.51 & 52.58 \\
        & MFP & 92.76 & -0.83 & 52.60 \\
        & GNN-RL & 93.49 & +0.10 & 54.00 \\
        & DepGraph & 93.64 & +0.11 & 61.00 \\ \cline{2-5}
        & \textbf{Ours} & 93.35 & +0.09 & \textbf{63.00} \\
    \midrule
    \multirow{7}*{ResNet-110} & Hrank & 94.23 & +0.73 & 41.20 \\
        & ABP & 93.95 & +0.32 & 46.20 \\
        & AGMC & 93.08 & -0.60 & 50.00 \\
        & GNN-RL & 94.31 & +0.63 & 52.00 \\
        & CHIP & 94.44 & +0.94 & 52.10 \\
        & MFP & 93.31 & -0.37 & 52.30 \\ \cline{2-5}
        & \textbf{Ours} & 93.91 & +0.41 & \textbf{55.00} \\
    \midrule
    \multirow{6}*{VGG-19} & SOSP & 73.11 & -0.34 & 51.61 \\
        & SCP & 72.15 & -0.41 & 61.94 \\
        & ProsPr & 72.29 & -0.21 & 80.00 \\
        & GREG & 67.55 & -6.47 & 88.69 \\
        & DepGraph & 70.39 & -3.11 & 88.79 \\ \cline{2-5}
        & \textbf{Ours} & 67.97 & -2.38 & \textbf{90.00} \\
    \bottomrule
    \end{tabular}
    
    \label{tab:cifar10_100}
    \vspace{-5mm}
\end{table}

\begin{table}
    \centering
    \caption{Pruning results on ImageNet-1K.}
    \setlength{\tabcolsep}{2.4mm}  
    \vspace{-6pt}
    
    \footnotesize
    \begin{tabular}{c|c|c|c|c}
    \toprule
    \textbf{DNN} & \textbf{Method} & \textbf{P. Acc.} & \textbf{$\Delta$ Acc.} & \textbf{FLOPs $\downarrow$} \\
    \midrule
    \multirow{6}*{MobileNet-v1} & SMCP & 71.00 & -1.60 & 37.43\\
        & DRLP & 70.60 & -0.30 & 50.00 \\
        & RollBack & 49.34 & - & 50.00 \\
        & AGMC & 69.40 & -1.20 & 60.00 \\
        & GNN-RL & 69.50 & -1.40 & 60.00 \\ \cline{2-5}
        & \textbf{Ours} & 69.85 & -1.29 & \textbf{70.00} \\
    \midrule
    \multirow{7}*{MobileNet-v2} & NPPM & 72.02 & +0.02 & 29.70 \\
        & RollBack & 52.86 & - & 40.00 \\
        & GNN-RL & 70.04 & -1.83 & 42.00 \\
        & GFP & 69.16 & -6.58 & 50.00 \\
        & DepGraph & 68.46 & -3.41 & 54.55 \\
        & DNCP & 66.50 & -5.80 & 67.67 \\ \cline{2-5}
        & \textbf{Ours} & 67.65 & -4.22 & \textbf{75.00} \\
    \midrule
    \multirow{6}*{ResNet-50} & GFP & 76.42 & -0.37 & 50.11 \\
        & SOSP & 74.39 & -1.76 & 51.00 \\
        & DepGraph & 75.83 & -0.32 & 51.82 \\
        & GNN-RL & 74.28 & -1.82 & 53.00 \\
        & MFP & 74.13 & -2.02 & 53.50 \\ \cline{2-5}
        & \textbf{Ours} & 74.03 & -2.10 & \textbf{65.00} \\
    \midrule
    \multirow{6}*{VGG-16} & NM & 61.18 & -12.18 & 38.41 \\
        & CC & 68.81 & -2.78 & 52.39 \\
        & GNN-RL & 70.99 & +0.49 & 80.00 \\
        & AMC & 69.10 & -1.40 & 80.00 \\
        & AGMC & 70.35 & -0.15 & 80.00 \\ \cline{2-5}
        & \textbf{Ours} & 71.47 & -0.12 & \textbf{82.00} \\
    \midrule
    \multirow{3}*{ViT-B/16} & CP-ViT & 77.36 & -0.55 & 33.52 \\
        & DepGraph & 79.17 & -1.90 & 40.90 \\ \cline{2-5}
        & \textbf{Ours} & 79.22 & -1.85 & \textbf{45.00} \\
    \bottomrule
    \end{tabular}
    
    \label{tab:imagenet}
    \vspace{-1mm}
\end{table}

\subsection{Results and Analysis}

{\bf Performance.} Table \ref{tab:cifar10_100} and \ref{tab:imagenet} present the results of our \model\ compared with baselines on two datasets. We can see that our method achieves the best results on compression ratio (FLOPs $\downarrow$) with an relative smaller accuracy variety ($\Delta$ Acc.), reaching the state-of-the-art. 
After pruning, the FLOPs of the models significantly decreased, while the inference accuracy could be restored to the level of the initial models after fine-tuning. This effectively maintains model performance while compressing the model, meaning that only the truly useful parameters are retained. 

Dig deeper, for all pruned DNNs, DNNs with simpler architecture can achieve higher pruning ratios, such as VGG-19 with 90\% and VGG-16 with 82\%. In contrast, DNNs containing more complex structures (e.g. residual connection and transformer encoder) will get more accuracy loss if we apply a higher pruning ratio, for instance, ViT-B/16 with 45\% pruning ratio, ResNet-50 and ResNet-110 with 55\% pruning ratio. In addition, our method achieves a reduction in FLOPs of nearly 18\% compared to the second-best auto-pruning method, GNN-RL, on ResNet-50, with the accuracy loss across all pruned models remaining under 3\%.

Moreover, we observed that for some models (such as ResNet-110 and MobileNet-v1), the accuracy after pruning even exceeded that of the original models. Upon analysis, we believe that the initial pre-trained models were not sufficiently trained, leading to improved inference accuracy after fine-tuning, surpassing the performance of the initial models.

Since some DNNs have similar architectures (e.g., ResNet-32, ResNet-56, and ResNet-110 all have residual structures but differ in network depth), we select ResNet-110, VGG-19, MobileNet-v1, and ViT-B/16 for the following experiments.

\pSpace
\noindent{\bf Latency Comparison.} We also test the inference latency of the pruned models and compare them with the original models, as shown in Table \ref{tab:latency}. We can see that our method significantly reduces the MACs, thereby lowering the inference latency while maintaining comparable or even improved accuracy. For example, in ResNet-110, the MACs were reduced from 0.26 G to 0.1 G, with a corresponding decrease in latency from 5.7 ms to 5.3 ms. Notably, the accuracy even slightly improved from 93.50\% to 93.91\%, indicating that the pruning method not only reduces the model size but also retains and potentially enhances its performance. 
For VGG-19, the MACs were drastically reduced from 0.4 G to 52.36 M, leading to a reduction in latency from 5.3 ms to 4.1 ms (nearly 29\% speed improvement). Although the accuracy slightly dropped from 70.35\% to 67.97\% (decreased by 2\%), this trade-off may be acceptable considering the significant improvement in inference speed and efficiency. 

\begin{table}
    \centering
    \caption{Inference latency of different DNNs.}
    \label{tab:latency}
    \vspace{-6pt}
    \setlength{\tabcolsep}{2.4mm} 
    \footnotesize
    \begin{tabular}{c|c|c|c|c}
    \toprule
    \textbf{DNN} & \textbf{Type} & \textbf{MACs} & \textbf{Acc.} & \textbf{Latency} \\
    \midrule
    \multirow{2}*{ResNet-110} & Original & 0.26 G & 93.50 & 5.7 ms \\
        & Pruned & 0.10 G & 93.91 & 5.3 ms \\
    \midrule
    \multirow{2}*{VGG-19} & Original & 0.40 G & 70.35 & 5.3 ms \\
        & Pruned & 52.36 M & 67.97 & 4.1 ms \\
    \midrule
    \multirow{2}*{MobileNet-v1} & Original & 0.59 G & 71.14 & 6.5 ms \\
        & Pruned & 0.24 G & 69.85 & 6.0 ms \\
    \midrule
    \multirow{2}*{ViT-B/16} & Original & 17.61 G & 81.07 & 5.4 ms \\
        & Pruned & 10.11 G & 79.22 & 5.1 ms \\
    \bottomrule
    \end{tabular}
\end{table}

\begin{figure}
    \centering
    
    \hspace{-3.2mm}
    \subcaptionbox{Acc. \wrt. number of patterns}{
        \label{fig:pattern_num_vary}
        \includegraphics[width=0.49\linewidth]{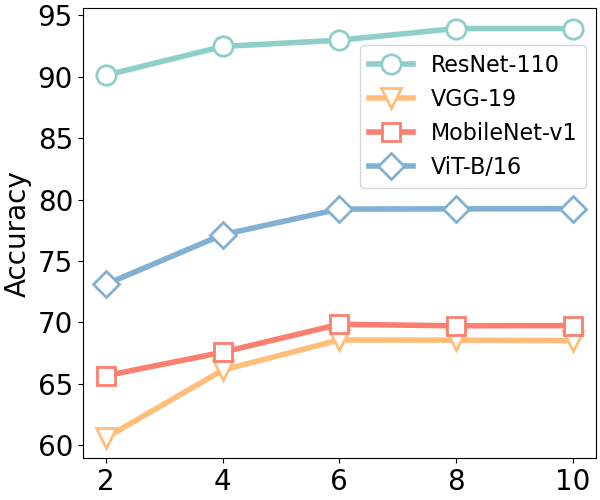}
    }
    \hspace{-2.82mm}
    \subcaptionbox{Acc. \wrt. node embedding}{
        \label{fig:embedding_vary}
        \includegraphics[width=0.49\linewidth]{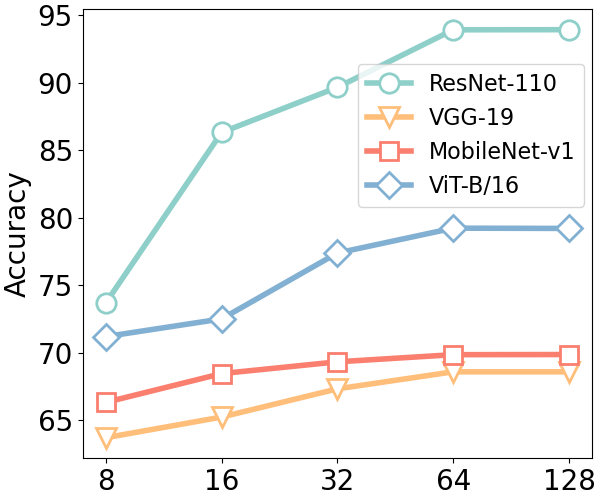}
    }
    \\
    \vspace{2mm}
    \hspace{-3.5mm}
    \subcaptionbox{Acc. \wrt. pruning ratio}{
        \label{fig:pruning_ratio_vary}
        \includegraphics[width=1.0\linewidth]{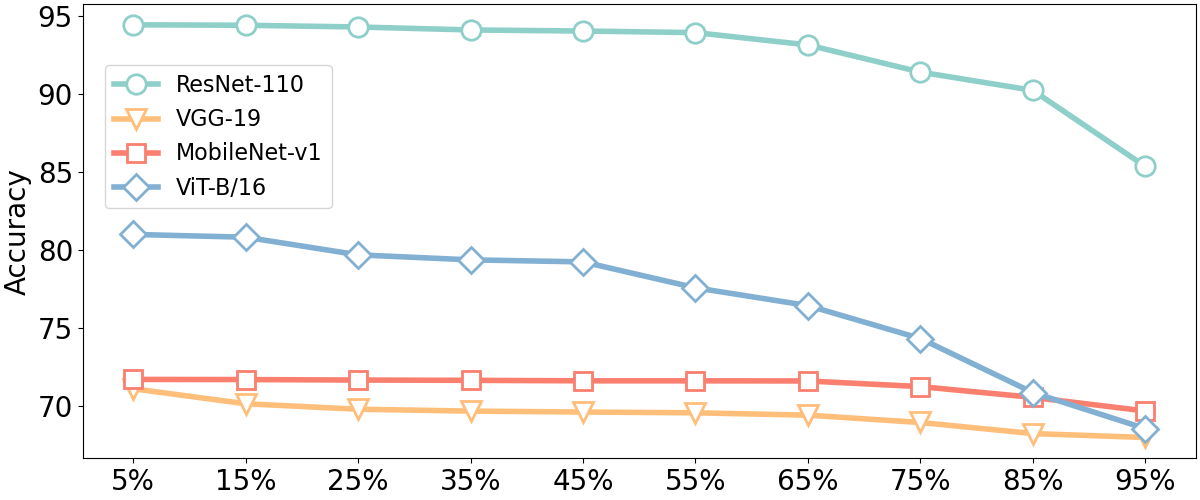}
    }
    \vspace{-2mm}
    
    \caption{Hyper parameter impact of patterns, node embedding size and pruning ratio.}
    \label{fig:hyper_param}
    \vspace{-1mm}
\end{figure}

\pSpace
\noindent{\bf Parameter Sensitivity.} We test the impact of different hyperparameters on the accuracy of pruned models, as shown in \figurename~\ref{fig:hyper_param}. First, we evaluate the effect of the number of patterns on the accuracy of pruned models. From \figurename~\ref{fig:hyper_param} (a), we can observe that increasing the number of patterns improves the accuracy of the inference for all models. Meanwhile, when the number of patterns reaches a certain threshold, the accuracy gains steadily. In particular, when the number of patterns is 2, it represents structured pruning, where the entire operator is either retained or pruned. Once the number of patterns exceeds a certain limit (e.g., 10), the granularity becomes finer, which can be regarded as unstructured pruning. Therefore, to balance the inference accuracy and compression rate, we use 6 patterns for pruning.
Next, we test the effect of the node embedding size for information aggregation by the graph encoder on the accuracy of pruned models, as shown in \figurename~\ref{fig:hyper_param} (b). We can see that ResNet-110 is more sensitive to node embedding size, likely due to its deeper and more complex network structure.
Lastly, we evaluate the changes in inference accuracy under different pruning ratios for each model, as depicted in Figure \ref{fig:hyper_param} (c). As expected, with increasing pruning ratios, the inference accuracy declines. However, different models exhibit varying sensitivity to pruning rates. For example, complex network architectures like ResNet-110 are more affected by changes in pruning ratio. In conclusion, the accuracy loss of our method is relatively stable with the increase of the pruning ratio.

\begin{figure}
    \centering
    \includegraphics[width=1.0\linewidth]{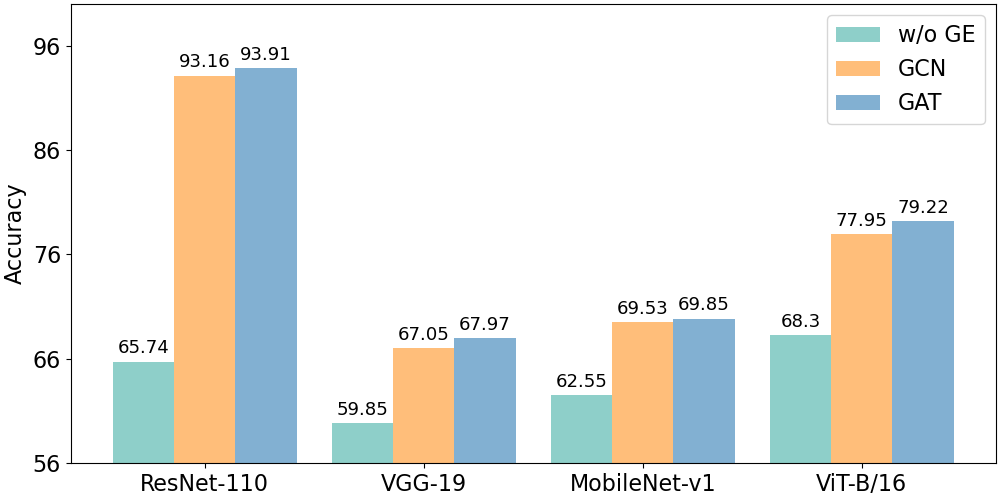}
    \caption{The effect of different DNN graph encoder.}
    \label{fig:graph_encoder}
    \vspace{-2mm}
\end{figure}

\begin{figure}[h]
    \centering
    \includegraphics[width=1.0\linewidth]{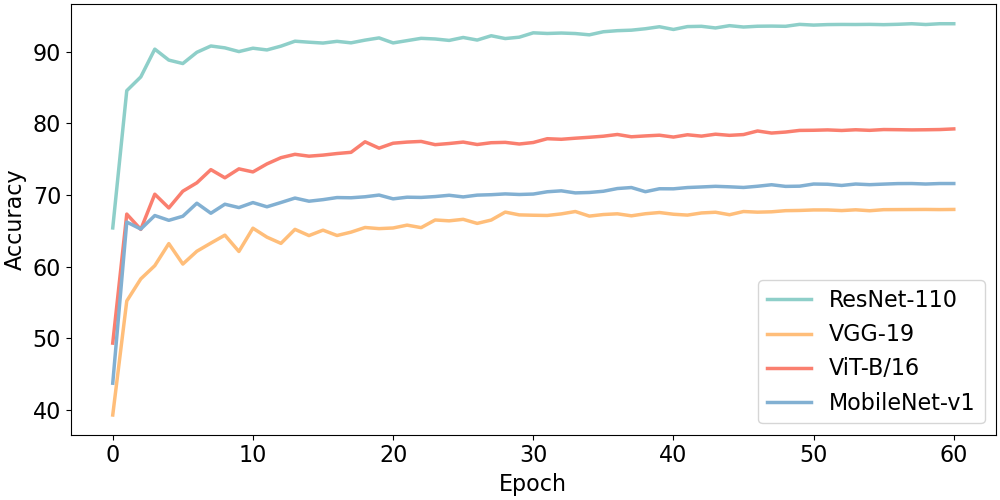}
    \caption{The efficiency of inference accuracy recovery.}
    \label{fig:accuracy_recovery}
    \vspace{-5mm}
\end{figure}
\noindent{\bf Efficiency of Accuracy Recovery.} \figurename~\ref{fig:accuracy_recovery} illustrates the efficiency of inference accuracy recovery after model pruning across different DNN architectures. As we can see, the models pruned by our method can rapidly recover from accuracy loss, demonstrating that the models pruned by our methods can be deployed with a lower cost. 

\subsection{Ablation Study}
We validate the effectiveness of incorporating graph encoders, including GCN and GAT, in the pruning process. We evaluate this on ResNet-110, VGG-19, MobileNet-v1, and ViT-B/16, as shown in \figurename~\ref{fig:graph_encoder}. The results demonstrate a significant improvement in the accuracy when using GCN or GAT to represent the DNN compared to the method (\textit{w/o GE}) without using graph encoders. For instance, the pruned ResNet-110's accuracy increased from 65.74\% without a graph encoder to 93.16\% with GCN and 93.91\% with GAT, highlighting the substantial advantage of using graph encoders, especially in deep networks like ResNet-110. Moreover, the results obtained with the GAT encoder were slightly better than those with the GCN encoder.
Overall, integrating graph encoders effectively captures the topological and pattern information within the deep neural network structure, leading to better performance after pruning.

\section{Conclusion}
We introduced AutoSculpt, a pattern-based model auto-pruning framework that uses GNN and DRL to efficiently compress DNNs. AutoSculpt represented DNNs as graphs and applied compute-friendly pruning patterns, allowing it to find better pruning strategies that work well across different DNN architectures and with standard inference engines. Experimental results showed that AutoSculpt achieved high compression rates with minimal accuracy loss, making it a strong solution for deploying DNNs on devices with limited resources. Future work could focus on improving pattern generation methods and expanding the framework to work with a wider range of DNN types, making it even more flexible and effective.

\clearpage
{
    \small
    \bibliographystyle{ieeenat_fullname}
    \bibliography{main}
}


\end{document}